\documentclass[letterpaper, 10 pt, journal, twoside]{ieeetran}
\usepackage[numbers]{natbib}
\usepackage{multicol}
\usepackage[table]{xcolor}
\usepackage[bookmarks=true]{hyperref}
\usepackage{amsmath} % assumes amsmath package installed
\usepackage{amssymb}  % assumes amsmath package installed
\usepackage{bm}
\usepackage{graphicx}
\usepackage{tabularx}
\usepackage{gensymb}
\usepackage{subcaption}
\usepackage{caption}
\usepackage{booktabs}
\usepackage{multirow}
\usepackage{makecell}
\usepackage[dvipsnames]{xcolor}

\renewcommand{\vec}[1]{\bm{#1}}
\newcommand{\update}[1]{#1}
\newcommand{\updatefin}[1]{#1}

\DeclareMathOperator*{\argmin}{arg\,min}

\definecolor{predpose}{RGB}{184, 183, 89}
\definecolor{gtpose}{RGB}{83, 123, 42}
\definecolor{predcontact}{RGB}{122, 5, 114}
\definecolor{gtcontact}{RGB}{250, 76, 80}

\newcommand{\orest}{\vec{o}_{r}}
\newcommand{\ot}{\vec{o}_t}
\newcommand{\gt}{\vec{g}_t}
\newcommand{\dispt}{\vec{\delta}_t}
\newcommand{\ctct}{\vec{c}_t}
\newcommand{\fort}{\vec{f}_t}
\newcommand{\wrencht}{\vec{w}_t}
\newcommand{\meshobj}{\mathcal{M}_o}
\newcommand{\meshenv}{\mathcal{M}_e}
\newcommand{\best}{\cellcolor{gray!25}}

\begin{document}

% paper title
\title{
Simultaneous Extrinsic Contact and In-Hand Pose Estimation via Distributed Tactile Sensing
}

\author{Mark Van der Merwe$^{1}$, Kei Ota${^2}$, Dmitry Berenson$^{1}$, Nima Fazeli$^{1}$, and Devesh K. Jha$^{2}$

%\thanks{Manuscript received: August 12, 2025; Revised November 13, 2025; Accepted December 15, 2025.}%Use only for final RAL version
\thanks{%This paper was recommended for publication by Editor Julia Borras Sol upon evaluation of the Associate Editor and Reviewers’ comments.
This work was supported in part by the Office of Naval Research Grant N00014-24-1-2036 and NSF grants IIS-2113401, IIS-2231607, and IIS-2220876.} %Use only for final RAL version
\thanks{$^{1}$Mark Van der Merwe, Dmitry Berenson, and Nima Fazeli are with Robotics Department, University of Michigan, USA
{\tt\footnotesize markvdm@umich.edu, dmitryb@umich.edf, nfz@umich.edu}}%
\thanks{$^{2}$Kei Ota and Devesh K. Jha are with Mitsubishi Electric Research Laboratories, USA {\tt\footnotesize ota@merl.com, jha@merl.com}}%
%\thanks{Digital Object Identifier (DOI): see top of this page.}
}

\maketitle

\begin{abstract}
Prehensile autonomous manipulation, such as peg insertion, tool use, or assembly, require precise in-hand \update{understanding} of the object pose and the extrinsic contacts made during interactions. Providing accurate estimation of pose and contacts is challenging. Tactile sensors can provide \update{local geometry at the sensor} and force information about the grasp, but the \update{locality of sensing} means resolving poses and contacts from tactile alone is often an ill-posed problem, \update{as multiple configurations can be consistent with the observations}. \update{Adding visual feedback can help resolve ambiguities, but can suffer from noise and occlusions. In this work, we propose a method that pairs local observations from sensing with the physical constraints of contact.} We propose a set of factors that ensure local consistency with tactile observations as well as enforcing physical plausibility, namely, that the \update{estimated} pose and contacts must respect the kinematic and force constraints of quasi-static rigid body interactions. We formalize our problem as a factor graph, allowing for efficient estimation. \update{In our experiments, we demonstrate that our method outperforms existing geometric and contact-informed estimation pipelines, especially when only tactile information is available.} \updatefin{Video results can be found at \href{tacgraph.github.io}{tacgraph.github.io}.}
\end{abstract}

% Keywords appear just beneath the abstract. Use only for final RAL version. 

% \IEEEpeerreviewmaketitle

\section{Introduction}

\IEEEPARstart{P}{rehensile} manipulation tasks require precise reasoning over grasped object poses and contacts. Even small errors in object pose can prevent proper insertion of an object~\cite{yu2018contact} or yield an undesired placement~\cite{li2024stable}. Effective tool use similarly relies on precise application of contacts and forces~\cite{holladay2019tool}. We require systems capable of simultaneously estimating where a grasped object is and how it is in contact with the environment.

\update{Methods for object pose estimation largely rely on visual feedback~\cite{schmidt2014dart,pmlr-v87-tremblay18a}. Prehensile manipulation, however, often suffers visual occlusions due to the grasp and/or environment \update{and sensor noise can corrupt estimation results}. Distributed visuo-tactile sensors are a promising form of feedback for prehensile manipulation~\cite{alspach2019soft,yuan2017gelsight} that can complement visual sensing~\cite{suresh2024neuralfeels}. These sensors utilize a compliant material that deforms on contact and a camera to observe these deformations. This provides local geometric information where the sensor contacts the object. While these geometric observations can resolve in-hand pose for small or highly featured objects~\cite{li2014gelsight}, for many objects, the local nature of these observations still results in significant ambiguities~\cite{bauza2023tac2pose} and must be paired with visual feedback for better convergence~\cite{bauza2024simple}.}

The compliance of the visuo-tactile sensor can also provide a signal of extrinsic contact, between the grasped object and the environment. How the sensor deforms when an extrinsic contact is made is indicative of the force experienced in the grasp~\cite{suh2022seed}, which in turn can be used as a signal to infer extrinsic contacts~\cite{sangwoon2023,sangwoon2022ecs}. Precisely identifying contact is still challenging, however, as multiple contact points can yield near-identical force profiles in the grasp.

\begin{figure}
    \centering
    \includegraphics[width=\linewidth]{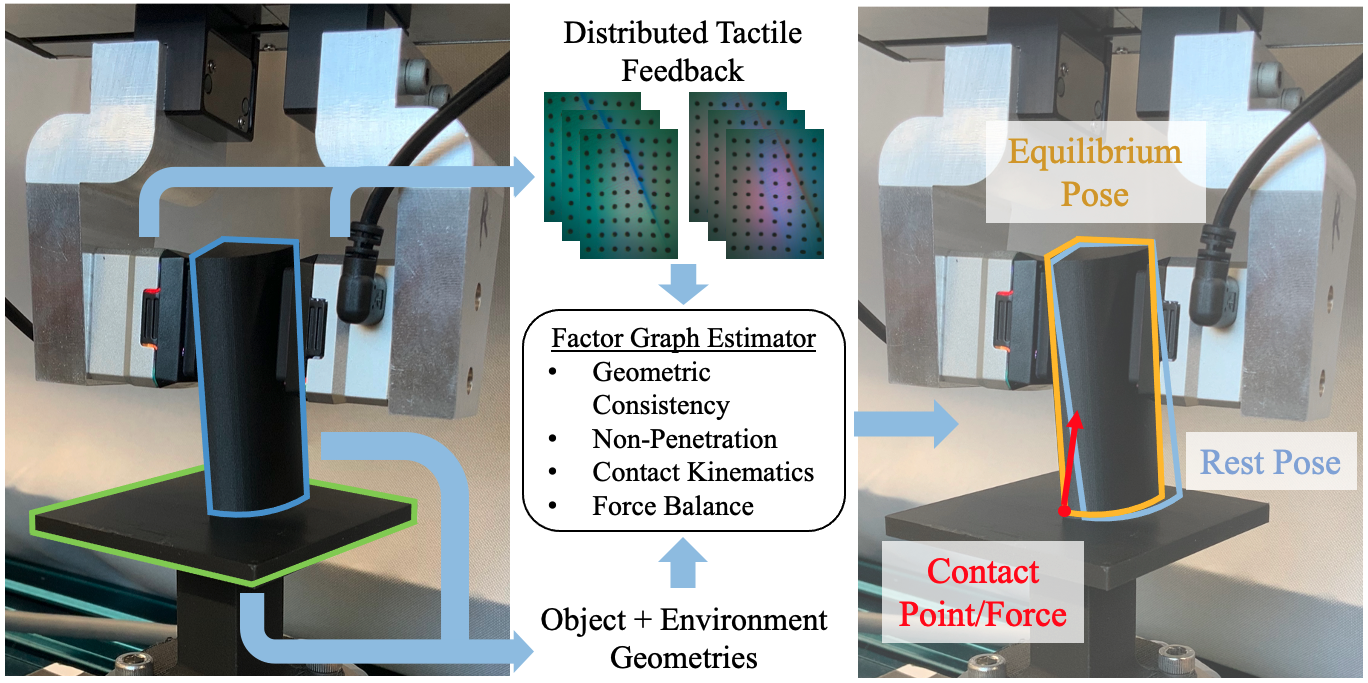}
    \caption{We propose \textit{TacGraph}, an estimator that exploits geometric consistency, force balance, non-penetration, and contact kinematics to jointly estimate the object pose and extrinsic contacts.}
    \label{fig:intro-figure}
\end{figure}

\update{In this work, we investigate how \textit{jointly} estimating the object pose and extrinsic contacts can help resolve ambiguities and address noise. Intuitively, these two are tightly coupled, as the object pose determines which contacts are possible, and conversely, where contact is being made constrains object poses.} We enforce non-penetration with the (assumed known) environment and ensure that the estimated extrinsic contact is kinematically feasible (i.e., lies on the surface of the object and environment) \update{and yields forces consistent with tactile feedback}. The result is a system that enforces consistency across multiple forms of feedback (geometry and forces) while enforcing physical realism (see Fig.~\ref{fig:intro-figure}). This reduces the space of acceptable solutions, resulting in accurate estimation.

Our contributions are as follows:
\begin{itemize}
    \item A set of \textit{object agnostic} models for extracting geometric and force signals for our estimator from tactile feedback.

    \item A factor graph method for simultaneous extrinsic contact and in-hand pose estimation, \emph{TacGraph}, which jointly enforces geometric consistency, force balance, non-penetration, and contact kinematics.

    \item Demonstration of our method on a physical system and comparison to baselines.
\end{itemize}
\section{Related Work}

\subsection{Extrinsic Contact Estimation}

Accurate recovery of extrinsic contacts is challenging due to the broad nature of contacts possible and the indirect and partial sensing available. Kim et al.~\cite{pmlr-v229-kim23b} learns to predict image contact masks directly from visual feedback. Other methods learn to predict line or patch geometries from point cloud and Force/Torque feedback~\cite{Merwe-RSS-23,pmlr-v205-merwe23a}. Higuera et al.~\cite{carolinancf} and Ota et al~\cite{ota2023tactile} predict extrinsic contact patches based on distributed visuo-tactile feedback. Lee et al.\cite{lee2025vitascope} predicts extrinsic contact patch along with in-hand pose. \update{These methods rely on expensive object-specific training, unlike our method which trains only object-agnostic components.}

Kim et al.~\cite{sangwoon2022ecs} and Kim et al.~\cite{sangwoon2023} utilize visuo-tactile feedback to estimate an extrinsic contact. Without knowledge of object geometry, their method utilizes the in-hand displacements observed from tactile sensing and active exploratory interactions to derive contact line or point. \update{However, this work does not consider object pose.}

\subsection{Prehensile Object Pose Estimation}

Tac2Pose~\cite{bauza2023tac2pose} learns to perform object pose estimation via an object-specific tactile model that returns pose distributions consistent with observed tactile feedback. Follow up work fuses incorporates visual feedback to further resolve ambiguity~\cite{bauza2024simple}. Dikhale et al.~\cite{9682507} fuse tactile and visual feedback to learn to predict object poses directly, akin to purely visual based tracking models~\cite{pmlr-v87-tremblay18a}.

\update{Several works investigate model-based estimation with combined visual and tactile feedback. Several works jointly reconstruct and estimate pose from visual and tactile geometric feedback~\cite{murali2025shared, suresh2024neuralfeels}. Zhong et al.~\cite{Zhong-RSS-23} extends model-based pose estimation to include free-space non-penetration constraints as well as tactile point clouds, but does not consider contact consistency of any form.}

\subsection{Joint Prehensile Pose and Extrinsic Contact Estimation}

\update{Bronars et al.~\cite{bronars2024texterity} extends Tac2Pose~\cite{bauza2023tac2pose} to include additional geometric contact constraints, however, it does not utilize the wrench consistency and relies on the object-specific models of Tac2Pose.} 

SCOPE~\cite{sipos2022simultaneous} and Multi-Scope~\cite{Sipos-RSS-23} propose a model-based approach that utilizes Force/Torque feedback on the robot and environment to simultaneously estimate object and environment pose and extrinsic contacts, utilizing the physical constraints of contact. \update{However, assuming F/T sensing on the environment is often unrealistic and their approach does not incorporate geometric consistency.}
\section{Problem Statement}\label{sec:problem}

Our goal is to estimate the pose of a grasped object and the extrinsic contact that the grasped object makes with a known environment. We make the following assumptions:

\begin{itemize}
    \item The object is rigid and has a known geometry $\mathcal{M}_o$.
    \item The environment is rigid, has a known geometry $\mathcal{M}_e$, and is static.
    \item The grasp is elastic, meaning it complies due to external contact, but the object returns to the same location when the extrinsic contact is removed.
    \item We assume that the extrinsic contact can be described as a single summary contact point and force, and does not induce a torque.
\end{itemize}

We utilize feedback from a pair of Gelsight tactile sensors~\cite{li2014gelsight}, located at the grasp, and robot proprioception. \update{For some experiments, we additionally provide visual feedback from an external camera.} Our inputs are:

\begin{itemize}
    \item (Optional) $\vec{P}^{V} \in \mathbb{R}^{N_V \times 3}$ - initial partial point cloud of object from external camera.
    \item $\vec{g}_t \in SE(3)$ - gripper pose at time $t$.
    \item $\vec{I}^L_t,\vec{I}^R_t\in \mathbb{R}^{H \times W \times 3}$ - Gelsight tactile images from the left and right gripper fingers at time $t$.
    \item $\mathcal{M}_o, \mathcal{M}_e$ - triangle mesh geometry of the grasped object and environment.
\end{itemize}

Our goal is to estimate the object pose as well as the contact point and contact force. Our desired outputs are:

\begin{itemize}
    \item $\vec{o}_t \in SE(3)$ - object pose at time $t$.
    \item $\vec{c}_{t} \in \mathbb{R}^3$ - contact point at time $t$.
    \item $\vec{f}_t \in \mathbb{R}^3$ - contact force at $\ctct$ at time $t$.
\end{itemize}

% \begin{figure}
%     \centering
%     \includegraphics[width=0.7\linewidth]{figures/variables.png}
%     \caption{An overview of the variables we aim to estimate in our system, as well as some of the provided observations. Note, shown in 2D for visual clarity; all terms are estimated in 3D. $\orest$ is the pose of the object in grasp when no forces are applied, while $\ot$ is the pose considering contact forces from the environment.}
%     \label{fig:variables}
% \end{figure}

% In Fig.~\ref{fig:variables}, we show a 2D simplification of our system setup, showing the variables and how they relate.
\section{TacGraph: Factor Graph Based Estimation} \label{sec:method}

\begin{figure*}
    \centering
    \includegraphics[width=0.9\linewidth]{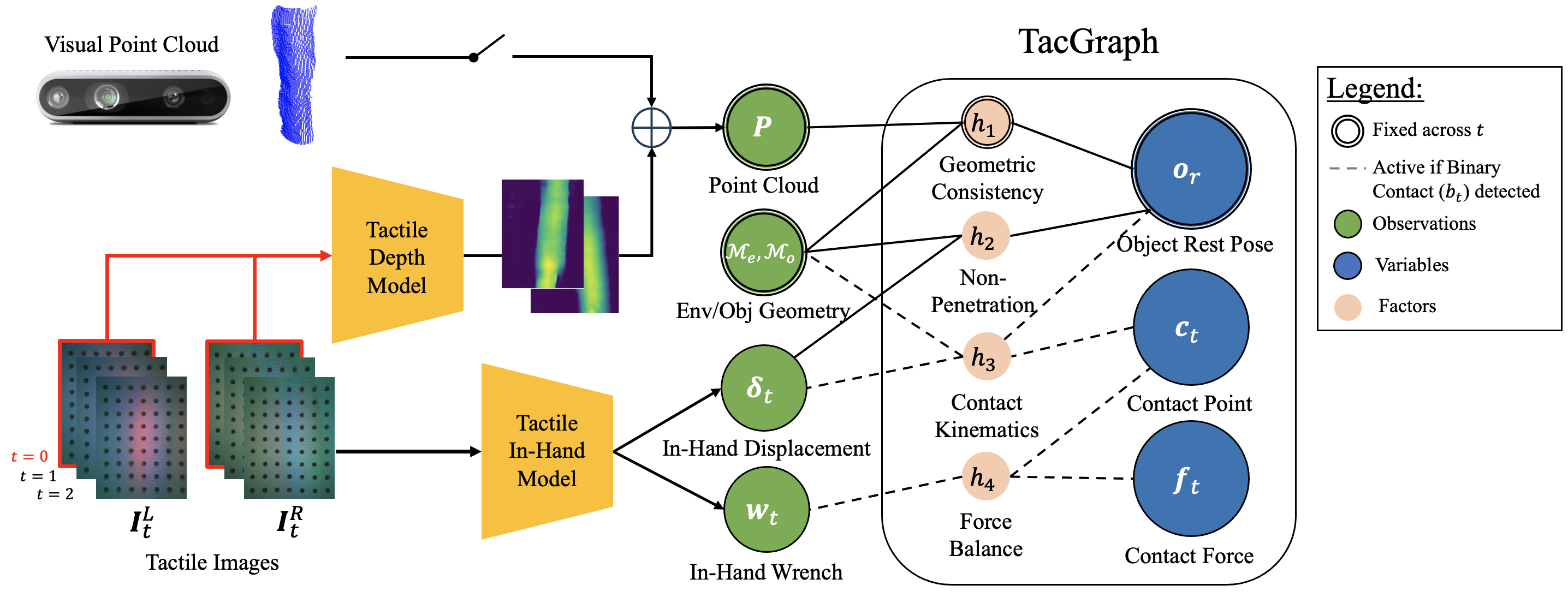}
    \caption{Overview of our proposed methodology. First, we propose a set of tactile models which process raw distributed tactile observations into geometric and force feedback terms. Second, these terms are utilized, along with known geometries and \update{optionally with visual feedback}, in a factor-graph based estimator, \textit{TacGraph}, which estimates the object pose and extrinsic contacts. Observations, variables, and factors that are fixed (non time-varying) are double circled. Factors only active when contact is detected are connected with dashed lines.}
    \label{fig:tacgraph}
\end{figure*}

We propose our method for simultaneously estimating in-hand object pose and extrinsic contacts from tactile feedback. First, we propose a set of object agnostic tactile models for extracting useful geometric and force signals to be used downstream. Second, we propose TacGraph, our factor-graph based estimator. An overview of our method is shown in Fig.~\ref{fig:tacgraph}.

\subsection{Tactile Models} \label{sec:tactile_models}

Distributed visuo-tactile sensors provide several forms of feedback that are valuable for contact-rich prehensile manipulation. First, we gain local geometric information from the grasp. Second, the deformation allowed by the sensor upon an external contact provides information about a) the in-hand displacement of the object and b) the force applied on the object. We propose a set of \textit{object-agnostic} models that extracts these feedback terms from the raw tactile images.

We use the initial tactile observations (i.e., at $t=0$), to predict a tactile point cloud $\vec{P}^T\in \mathbb{R}^{N_T \times 3}$. We train a model that takes in a tactile image (either from the left or right sensor) and yields a depth image $D\in \mathbb{R}^{H\times W}$. We threshold the depth to determine which pixels are in contact and de-project the depths to yield our final point cloud $\vec{P}^T$.

As the grasped object makes contact with its environment, the compliance of the Gelsight sensor means the object moves in the grasp. In order to accurately recover the object pose, we must then reason about this displacement. We train an in-hand displacement model which takes in both tactile images and predicts $\vec{\delta}_t \in SE(3)$, the relative displacement of the object in-hand.

Finally, we wish to determine the in-hand wrench experienced at the grasp, which is helpful for contact localization and for resolving the contact force~\cite{manuelli2016,sipos2022simultaneous}. We train an additional model that takes in both tactile images and predicts $\vec{w}_t \in \mathbb{R}^6$, the wrench experience at the grasp. Additionally, we can utilize our predicted wrench to estimate whether the system is in contact at time $t$: $b_t = ||\vec{w}_t||_\Sigma > \epsilon$.

\subsection{TacGraph}

Our goal is to incorporate our tactile feedback terms along with the physical constraints of contact to determine the \textit{maximum a posteriori} (MAP) estimate of our variables. As stated in Sec.~\ref{sec:problem}, we aim to recover the object pose and extrinsic contact information at each time $t$.

We assume that the grasp is elastic, that is, the object moves in-hand due to sensor compliance upon the application of an external force, and returns to the same rest pose when the external force is removed. As described in Sec. $\ref{sec:tactile_models}$, we estimate this in-hand displacement $\dispt$ from the tactile feedback. As such, we perform a change of variables and infer $\ot$ by estimating a single \textit{rest pose} $\orest\in SE(3)$ and applying the predicted in-hand displacement at a given time $t$.
\begin{equation} \label{eq:object_pose}
    \ot = \gt{^g\dispt}{^g\orest}
\end{equation}
We use the prefix $g$ to indicate that the displacement and rest pose are both expressed in the gripper pose frame, but drop the frames from here on.

A natural way to describe our MAP estimation problem is as a Factor Graph, a bi-partite graph of variables and the factors which describe the relationship between variables~\cite{dellaert2017factor}. Assuming Gaussian noise models on the factors, the MAP estimation becomes a sum of nonlinear least-squares.
\begin{equation} \label{eq:map_inf}
    \orest^*, \vec{c}^*_{1:T}, \vec{f}^*_{1:T} = \argmin_{\orest, \vec{c}_{1:T}, \vec{f}_{1:T}} H(\orest, \vec{c}_{1:T}, \vec{f}_{1:T})
\end{equation}
\begin{equation} \label{eq:tacgraph_cost}
    \begin{aligned}
        H(\cdot) =& ||h_1(\orest)||_{\Sigma_1}^2 + \sum_{t=1}^T \{||h_2(\orest)||_{\Sigma_2}^2 + \\
    & \mathbf{1}[b_t] (||h_3(\orest, \ctct)||_{\Sigma_3}^2 + ||h_4(\ctct, \fort)||_{\Sigma_4}^2)\}
    \end{aligned}
\end{equation}
The structure of the factor graph is shown in Fig.~\ref{fig:tacgraph} in the box. Note, $b_t$ is an observation (see Sec.~\ref{sec:tactile_models}), not a variable being solved for, and is used to enable contact-specific factors only when the object is actively in contact. Each factor $h_i$ has an accompanying covariance matrix $\Sigma_i$ which is empirically selected. The resulting MAP problem can be solved efficiently using the iSAM2 solver~\cite{kaess2012isam2}.

\subsection{Factors}

\subsubsection{Geometric Consistency} \label{factor:tactile-point-cloud} Our first factor ensures the estimated rest pose $\orest$ is \update{consistent with the observed object point cloud at time $t=0$. This point cloud $\vec{P}$ includes the tactile point cloud $\vec{P}^T$ and, if available, a visual point cloud $\vec{P}^V$.} As we assume that the grasp is elastic, it is sufficient to ensure the rest pose is geometrically consistent, without enforcing at each time step.

We define our factor error function as the signed-distance value of each point of the observed point cloud to the surface of the object:
\begin{equation}
    h_1(\orest;\vec{P}, \meshobj) = \vec{S}
\end{equation}
Here, $\vec{S}$ is a vector containing the signed distance value of each point to the surface.
\begin{equation}
    \vec{S}_i = SDF(\orest^{-1} \vec{P}^T_i | \meshobj)
\end{equation}
$SDF$ is the signed distance value computed using the geometry $\meshobj$. We transform each point into the object frame using the estimated $\orest$. Both the SDF and the SDF gradients can be efficiently computed using a triangle mesh geometry~\cite{Zhong-RSS-23}.

\subsubsection{Non-Penetration} We ensure the estimated pose at each time-step does not yield penetration with the environment geometry. We apply Eq.~\ref{eq:object_pose} to get the estimated object pose, considering the observed gripper pose and in-hand displacement, as well as the current estimate of the rest pose.

Given the known geometries $\meshenv, \meshobj$ and the object pose, we then compute a penetration check. In practice, we approximate our object geometry for this factor as a point cloud $\vec{P}^{obj}\in\mathbb{R}^{N_P \times 3}$ by sampling points on the surface of the geometry $\meshobj$. Our factor error function then returns for each point the distance to the surface of the environment, if the point is inside of the environment geometry, thus indicating penetration.
\begin{equation}
    h_2(\orest|\meshobj,\meshenv,\gt,\dispt) = \vec{S}
\end{equation}
Once again $\vec{S}$ is a vector which contains the SDF terms when a point is in penetration.
\begin{equation}
    \vec{S}_i = \min(0, SDF(\gt \dispt \orest \vec{P}_i^{obj}|\meshenv))
\end{equation}
Each point is transformed into the world frame before evaluating the SDF against the environment geometry.

\subsubsection{Contact Kinematics} When we detect that contact has been made $b_t$, we add additional variables and factors to the graph at that time step to estimate the contact location and force. First, we address the kinematic constraints on the contact. Namely, the contact point $\ctct$ should lie on the surface of \textit{both} the environment and object geometries. This amounts to the following error function:
\begin{equation}
    h_3(\orest, \ctct; \meshenv, \meshobj) = \begin{bmatrix}
        SDF(\ctct | \meshenv) \\
        SDF((\gt \dispt \orest)^{-1} \ctct | \meshobj)
    \end{bmatrix}
\end{equation}
We again apply Eq.~\ref{eq:object_pose} to derive the object pose and invert to map the point to the object frame.

\subsubsection{Force Balance}\label{sec:wrench-consistency} In Sec.~\ref{sec:tactile_models}, we showed how we can recover the wrench experienced at the grasp $\wrencht$ from our tactile sensors. This in-hand wrench is the result of the application of an external contact force. As such, we add a factor which ensures that the in-hand wrench implied by the contact matches our $\wrencht$ observation.

To estimate the in-hand wrench applied by a contact force $\fort$ applied at $\ctct$, we apply the contact Jacobian to map the force from the contact point to the gripper frame.
\begin{equation}
    \hat{\vec{w}}_t = J(\gt^{-1}\ctct)\ 
    {\fort}
\end{equation}
We can then define our factor error function as the difference between the observed and predicted in-hand wrench.
\begin{equation}
    h_4(\ctct, \fort; \gt) = \hat{\vec{w}}_t - \wrencht
\end{equation}

\subsection{Inference} \label{sec:inference}

Solving for the MAP estimate with our proposed TacGraph enables us to find the most likely rest pose $\orest$ and contact points $\vec{c}_{1:T}$ and forces $\vec{f}_{1:T}$. We then apply Eq.~\ref{eq:object_pose} to resolve our object poses at each timestep $\vec{o}_{1:T}$.

\update{The non-linear nature of our system means that the minimization performed by iSAM2 in Eq.~\ref{eq:map_inf} is subject to local minimums. As such, we initialize a set of particles to represent possible rest poses $\{ \orest^1, \orest^2, \ldots, \orest^K \}$. We then perform our inference procedure on each rest pose particle separately, resulting in $K$ solutions. We utilize our factor graph cost in Eq.~\ref{eq:tacgraph_cost} to score each solution particle, and select our final solution accordingly:
\begin{equation}
    \orest^*, \vec{c}^*_{1:T}, \vec{f}^*_{1:T} = \argmin_{k} H(\orest^k, \vec{c}^k_{1:T}, \vec{f}^k_{1:T})
\end{equation}}

\update{We utilize Iterative Closest Point (ICP) to match our initial rest pose particles to the available point cloud $\vec{P}$. We apply a sampling heuristic based on likely grasp directions~\cite{bauza2023tac2pose} to ensure that the rest pose particles cover the space of possible initializations well.} 
\section{Implementation} \label{sec:implementation}

\begin{figure}
    \centering
    \includegraphics[width=0.7\linewidth]{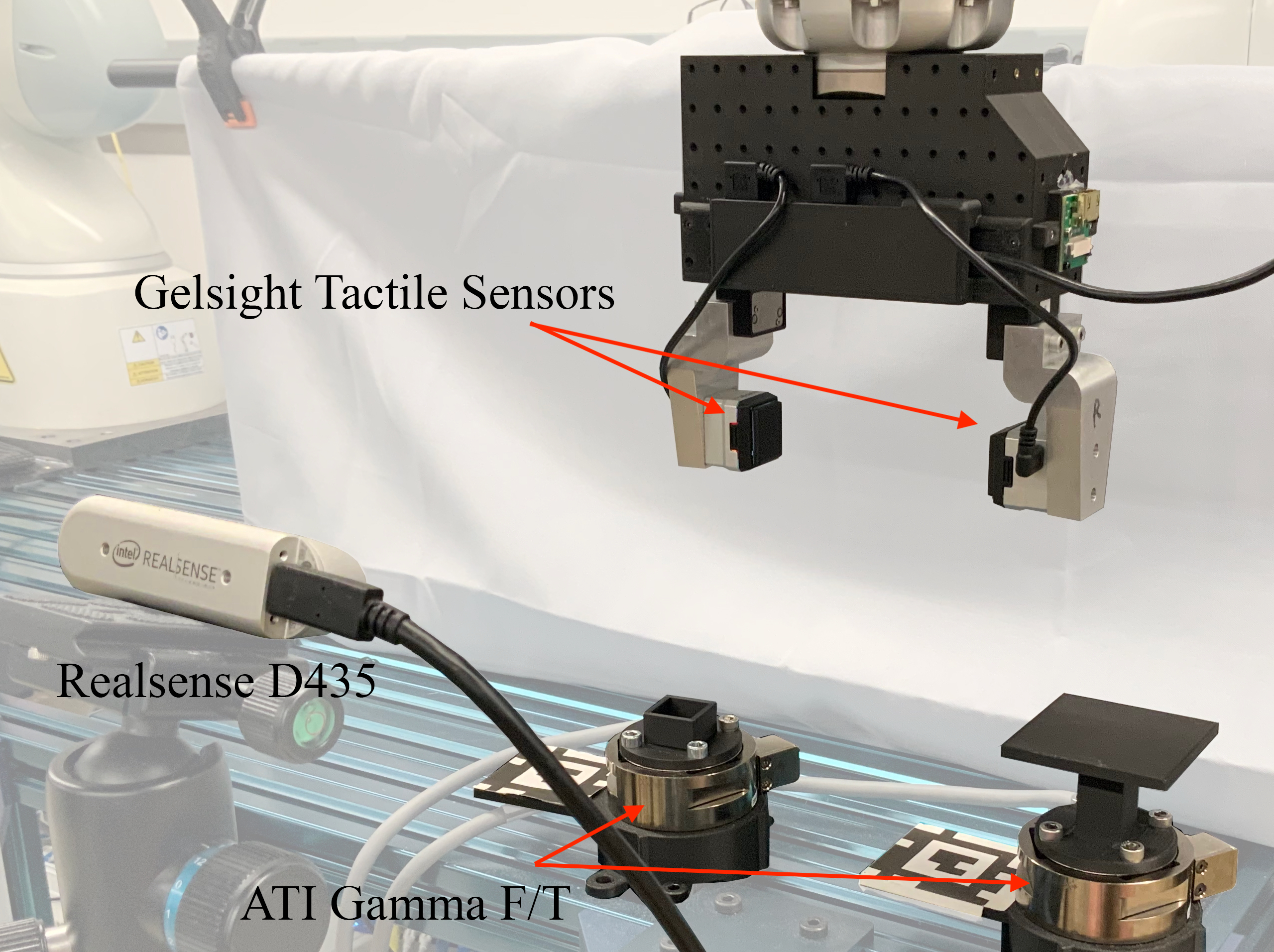}
    \caption{Our experimental setup. On the left ATI Gamma is an example object fixture. On the right ATI Gamma is the sensorized press surface we use for data collection/experiments.}
    \label{fig:exp_setup}
\end{figure}

\begin{figure}
    \centering
    \includegraphics[width=\linewidth]{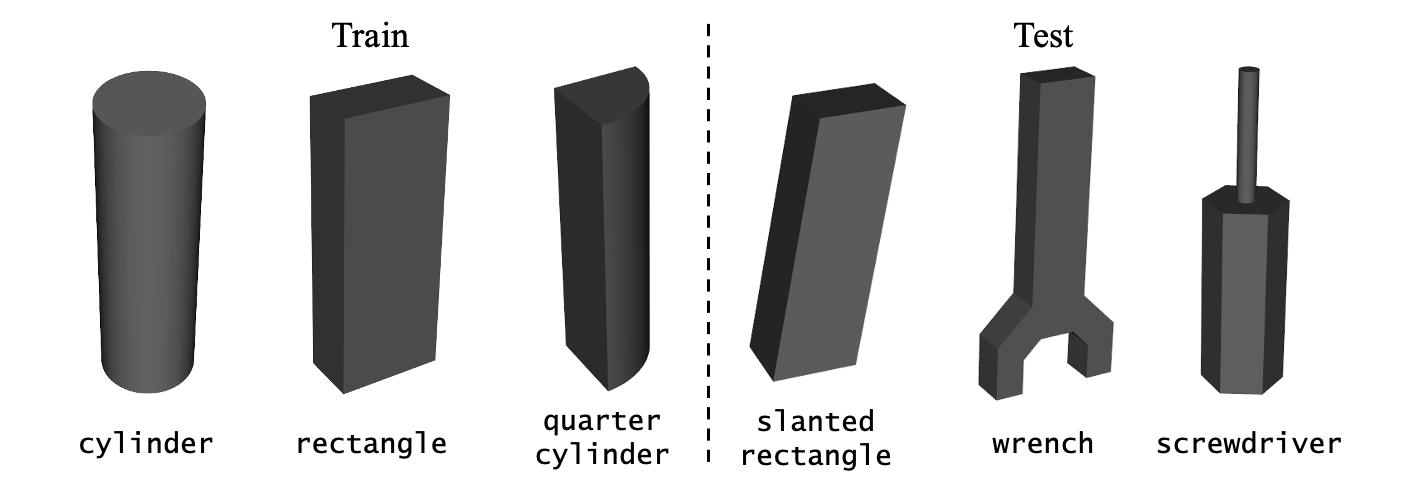}
    \caption{Train/Test objects used in our experiments.}
    \label{fig:objects}
\end{figure}

\begin{figure*}[t]
    \centering

    % LEFT COLUMN
    \begin{minipage}[t]{0.24\textwidth}
        \centering
        \vspace{8pt}

        \vfill

        \begin{subfigure}[t]{\linewidth} % <-- use full minipage width
            \vspace{0pt}
            \centering
            \includegraphics[width=0.65\linewidth]{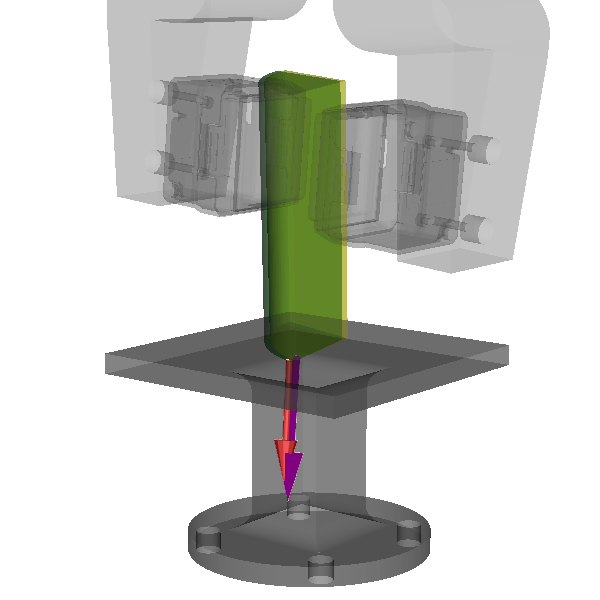}
            \caption{\texttt{quarter cylinder}}
        \end{subfigure}

        \vspace{0.5em}

        \begin{subfigure}[t]{\linewidth}
            \vspace{0pt}
            \centering
            \includegraphics[width=0.65\linewidth]{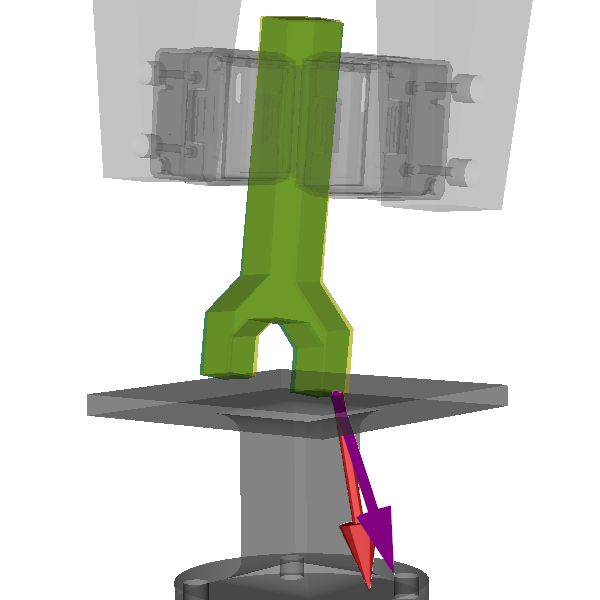}
            \caption{\texttt{wrench}}
        \end{subfigure}

        \vfill
    \end{minipage}
    \hfill
    % RIGHT COLUMN
    \begin{minipage}[t]{0.75\textwidth}
        \vspace{0pt} % <-- critical
        \centering

        \begin{subfigure}[t]{\linewidth}
            \vspace{0pt}
            \centering
            \includegraphics[width=0.9\linewidth]{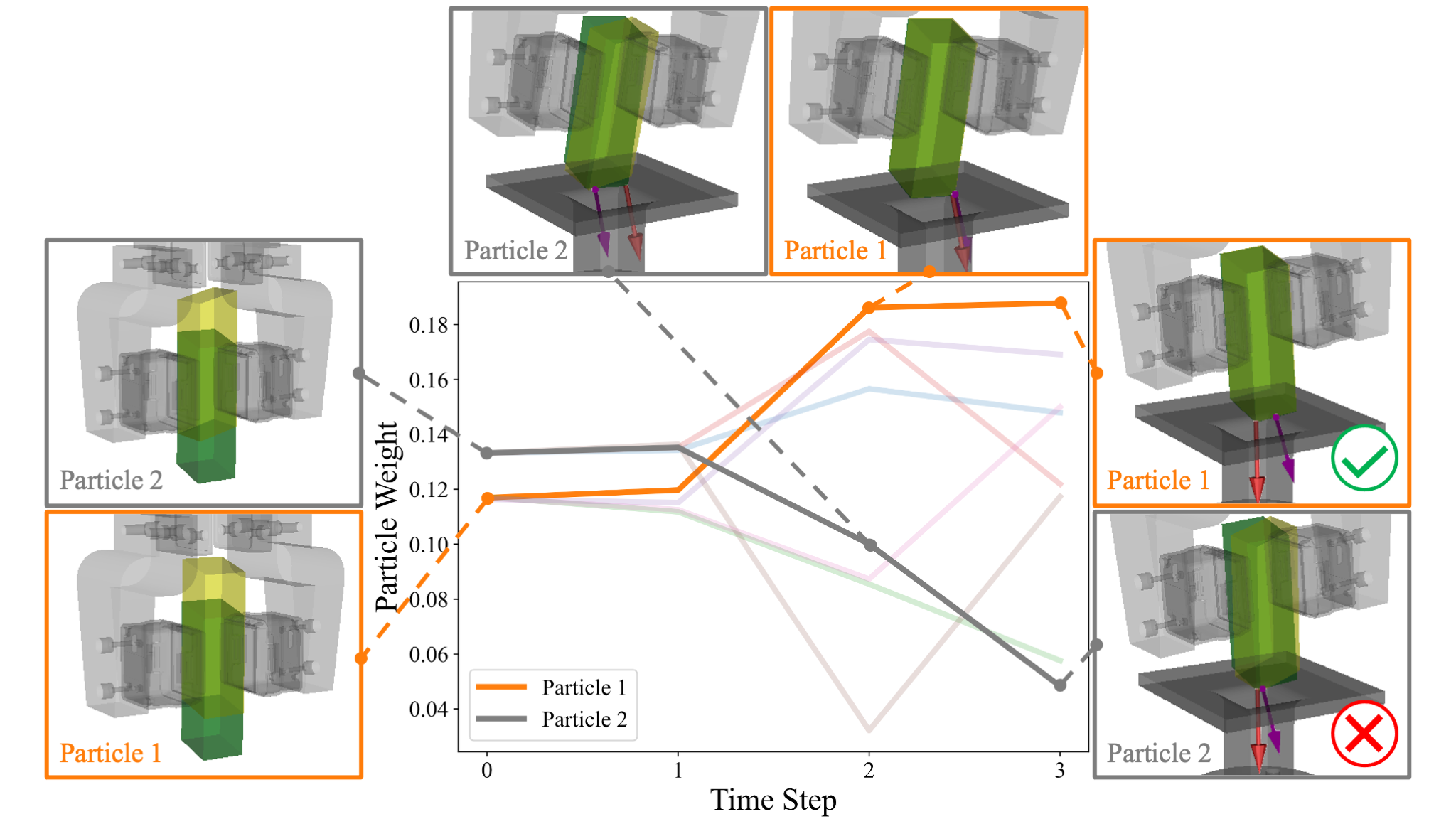}
            \caption{Particle Weight Progression}
            \label{fig:particles_qual}
        \end{subfigure}
    \end{minipage}
    \caption{We show comparison of \textcolor{predpose}{predicted} and \textcolor{gtpose}{ground truth} object pose, and \textcolor{predcontact}{predicted} and \textcolor{gtcontact}{ground truth} extrinsic contact. (a-b) final qualitative TacGraph estimates for two different objects. \updatefin{(c) progression of particle weights within TacGraph for a tactile-only inference. Two highlighted particles indicate how initial orientations can be filtered based on the contacts made to correctly select particle solutions.} Figure best viewed in color.}
    \label{fig:particle_qual}
\end{figure*}

\subsection{Experimental Setup}

Our experimental setup is shown in Fig.~\ref{fig:exp_setup}. We use a WSG-50 parallel jaw gripper attached to a KUKA LBR iiwa Med R820 robot. We attach a GelSight Mini sensor to each finger. We have an external Intel Realsense D435 sensor for visual feedback. \update{We utilize Segment Anything Model for visual point cloud segmentation~\cite{kirillov2023segment}.} Two ATI Gamma Force/Torque sensors are mounted in the scene and used to attach fixtures and environment geometries. We utilize these extrinsic F/T sensors \textit{only} for training our tactile models and evaluation; our method does not utilize these sensors' feedback. Our tactile models are implemented using Pytorch and our factor graph is implemented using GTSAM~\cite{gtsam}.

\subsection{Tactile Model Training}

We train our tactile models described in Sec.~\ref{sec:tactile_models} utilizing datasets collected on our physical system, using three different train objects (see Fig.~\ref{fig:objects}). All models are convolutional models trained via supervised learning.

\subsubsection{Tactile Depth} To generate tactile images with ground truth depth labels, we rigidly attach fixtures of known geometry to our environment and apply random grasps to the fixture, adjusting the position and orientation of the grasp. Then using the known fixture geometry, we render corresponding depth images. % Our depth model is a U-Net style convolutional neural network~\cite{ronneberger2015unet}, trained from scratch on our dataset. We found utilizing an RGB difference image between the grasped tactile images and the base tactile images with no object to work better than directly inputting the tactile image.

\subsubsection{In-Hand Displacement} To generate in-hand displacement labels, we rigidly attach fixtures of known geometry to our environment. Following Kim and Rodriguez~\cite{sangwoon2022ecs}, we randomly grasp the fixture, then apply small delta motions to the end effector. These delta motions become the in-hand displacement labels $\dispt$ associated with the tactile images.

% Rather than directly inputting the raw RGB images, we compute optical flow to extract a shear field representation~\cite{taylor2022gelslim, bogert2024built}, $\vec{U}^{H\times W \times 2}$ per sensor. We then input the vector field to a convolutional neural network, then predict displacement from a Multi-Layer Perceptron (MLP).

\subsubsection{In-Hand Wrench} To generate in-hand wrench labels, we utilize the tabletop fixture shown in Fig.~\ref{fig:exp_setup} and grasp objects, with randomized grasps, from a fixture before performing a random poke into the tabletop. We then use the environment mounted F/T sensor to get the wrench caused by the poke $\vec{w}_t^{e}$. We transform this wrench into the grasp frame, recovered from the robot proprioception, to label the wrench. % We again use the shear-field representation recovered from optical flow. We use the same convolution neural network, followed by a MLP to regress the wrench.

\section{Experiments}

% \begin{figure}
%     \centering
%     \includegraphics[width=\linewidth]{figures/insertion_particles_v2.png}
%     \caption{Example low-cost solution particles for a \texttt{slanted rectangle} TacGraph inference. Due to the partial nature of the initial point cloud, the particles at $t=0$ contain multiple matching orientations. After contact ($t=1$), only the correct orientation is consistent with the contact feedback.}
%     \label{fig:tacgraph_insert_particles}
% \end{figure}

\begin{table*}
    \centering
    \renewcommand{\arraystretch}{1.2}
    \begin{tabular}{c|c ccc ccc}
    \toprule
    & \multirow{2}{*}{Methods} & \multicolumn{3}{c}{Train} & \multicolumn{3}{c}{Test} \\
    \cmidrule(lr){3-5} \cmidrule(lr){6-8}
     & & \texttt{cylinder} & \texttt{rectangle} & \makecell{\texttt{quarter}\\ \texttt{cylinder}} & \makecell{\texttt{slanted} \\ \texttt{rectangle}} & \texttt{wrench} & \texttt{screwdriver} \\
    \hline

    \multirow{4}{*}{\rotatebox{90}{\makecell{Vision +\\Tactile}}}
    
     & ICP & 3.08 (0.65) & 3.68 (0.73) & 3.09 (1.08) & 4.86 (4.81) & 2.19 (0.76) & 3.33 (1.25) \\
     & CHSEL & 1.04 (0.36) & 1.97 (0.39) & \best{1.23 (0.51)} & 12.28 (5.11) & 1.66 (0.88) & 2.44 (0.83) \\
     & SCOPE (v1) & 5.96 (2.83) & 7.11 (2.70) & 16.50 (2.69) & 15.56 (4.36) & 5.86 (1.68) & 7.34 (2.53) \\
     & \updatefin{SCOPE (v2)} & \updatefin{4.77 (2.50)} & \updatefin{2.59 (1.22)} & \updatefin{14.04 (5.55)} & \updatefin{13.28 (4.79)} & \updatefin{3.45 (1.57)} & \updatefin{5.63 (3.11)} \\
     & TacGraph & \best{0.68 (0.23)} & \best{1.48 (0.14)} & 1.34 (0.34) & \best{1.05 (0.32)} & \best{0.92 (0.28)} & \best{1.62 (0.22)} \\

    \hline 

    \multirow{4}{*}{\rotatebox{90}{Tactile}}
     & ICP & 17.80 (13.86) & 17.15 (10.66) & 23.70 (7.58) & 20.10 (6.94) & 16.77 (6.03) & 12.32 (7.87) \\
     & CHSEL & 30.33 (12.01) & 28.55 (17.20) & 28.40 (6.86) & 20.71 (7.09) & 13.61 (8.16) & 29.42 (17.20) \\
     & SCOPE (v1) & 9.01 (6.49) & 12.26 (12.38) & 16.13 (10.33) & 18.44 (5.08) & 9.89 (7.39) & 8.32 (3.85) \\
     & \updatefin{SCOPE (v2)} & \updatefin{4.16 (1.00)} & \updatefin{4.37 (2.78)} & \updatefin{10.74 (5.37)} & \updatefin{12.90 (4.94)} & \updatefin{4.75 (3.77)} & \updatefin{11.03 (5.31)} \\
     & TacGraph & \best{2.96 (2.54)} & \best{1.29 (2.24)} & \best{8.45 (5.70)} & \best{7.58 (6.18)} & \best{0.78 (0.40)} & \best{1.54 (1.29)} \\
    
    \bottomrule
    \end{tabular}
    \caption{\update{Quantitative results for pose estimation. Mean and std. dev. of Averaged 3D Distance (ADD) in mm reported; best for each method in Vision+Tactile and Tactile regime highlighted.}}
    \label{tab:pose_est}
\end{table*}

\begin{table*}
    \centering
    \renewcommand{\arraystretch}{1.2}
    \begin{tabular}{c|c ccc ccc c}
    \toprule
    & \multirow{2}{*}{Methods} & \multicolumn{3}{c}{Train} & \multicolumn{3}{c}{Test} & \multirow{2}{*}{Overall} \\
    \cmidrule(lr){3-5} \cmidrule(lr){6-8}
     & & \texttt{cylinder} & \texttt{rectangle} & \makecell{\texttt{quarter}\\ \texttt{cylinder}} & \makecell{\texttt{slanted} \\ \texttt{rectangle}} & \texttt{wrench} & \texttt{screwdriver} & \\
    \hline

    \multirow{4}{*}{\rotatebox{90}{\makecell{Vision +\\Tactile}}}
     & ICP & 3.43 (0.80) & 4.77 (2.10) & 4.33 (2.48) & 5.30 (5.68) & 2.61 (1.34) & 4.38 (1.17) & 4.14 (2.92) \\
     & CHSEL & \best{1.84 (1.02)} & \best{3.15 (2.72)} & 2.79 (2.62) & 12.59 (9.44) & 2.44 (1.41) & \best{3.15 (0.55)} & 4.33 (5.62) \\
     & SCOPE (v1) & 5.43 (2.33) & 9.59 (8.45) & 6.33 (3.09) & 10.97 (4.51) & 4.84 (3.53) & 36.51 (14.33) & 12.28 (13.28) \\
     & \updatefin{SCOPE (v2)} & \updatefin{4.64 (2.48)} & \updatefin{8.71 (8.52)} & \updatefin{5.33 (3.38)} & \updatefin{9.48 (5.35)} & \updatefin{3.75 (1.87)} & \updatefin{27.10 (17.32)} & \updatefin{9.83 (11.59)} \\
     & TacGraph & 2.78 (1.31) & 4.79 (4.63) & \best{1.93 (1.06)} & \best{7.03 (6.11)} & \best{2.29 (1.13)} & 3.29 (1.20) & \best{3.68 (3.72)} \\

    \hline 

    \multirow{4}{*}{\rotatebox{90}{Tactile}}
    & ICP & 18.20 (13.49) & 17.38 (10.98) & 18.65 (8.92) & 20.71 (8.19) & 17.06 (5.54) & 13.20 (7.70) & 17.53 (9.75) \\
     & CHSEL & 25.16 (14.40) & 32.66 (14.68) & 22.54 (9.46) & 21.15 (10.17) & 13.54 (8.08) & 31.41 (18.30) & 24.41 (14.52) \\
     & SCOPE (v1) & 8.19 (8.31) & 15.49 (12.55) & 9.25 (12.25) & 16.36 (8.89) & 11.15 (8.39) & 36.74 (13.21) & 16.20 (14.49) \\
     & \updatefin{SCOPE (v2)} & \updatefin{4.73 (2.25)} & \updatefin{11.16 (10.65)} & \updatefin{\best{3.72 (2.59)}} & \updatefin{9.34 (5.32)} & \updatefin{6.98 (5.35)} & \updatefin{26.49 (13.63)} & \updatefin{10.40 (10.93)} \\
     & TacGraph & \best{3.29 (1.52)} & \best{2.60 (2.02)} & 3.82 (2.55) & \best{7.63 (7.03)} & \best 2.34 (0.85) & \best 2.64 (1.69) & \best 3.72 (3.78) \\
    
    \bottomrule
    \end{tabular}
    \caption{\update{Quantitative results for contact point estimation. Mean and std. dev. of distance to G.T. contact point in mm reported; best for each method in Vision+Tactile and Tactile regime highlighted.}}
    \label{tab:contact_est}
\end{table*}

\begin{table}
    \centering
    \renewcommand{\arraystretch}{1.2}
    \begin{tabular}{c cc}
    \toprule
    Methods & Vision + Tactile & Tactile \\
    \hline
    ICP & 0.68 (0.39) & 0.68 (0.39) \\
    CHSEL & 0.68 (0.39) & 0.68 (0.39) \\
    SCOPE (v1) & \best 0.61 (0.37) & 0.63 (0.44) \\
    \updatefin{SCOPE (v2)} & \updatefin{0.62 (0.37)} & \updatefin{\best 0.61 (0.41)} \\
    TacGraph & \best 0.61 (0.36) & \best 0.61 (0.36) \\
    \bottomrule
    \end{tabular}
    \caption{\updatefin{Summary Quantitative results for contact force estimation. Mean and std. dev. of difference to G.T. contact force in Newtons reported; best for each method in Vision+Tactile and Tactile regime highlighted.}}
    \label{tab:force_est}
\end{table}

\subsection{Baselines}

\update{\begin{itemize}
    \item \textbf{ICP:} Iterative Closest Point enforces geometric consistency in the object pose. To estimate a contact point and force, we use our tactile models to detect when contact occurs as well as the force. When in contact, we select the point on the object closest to the environment as our contact point, assigning the predicted force from the tactile model.
    \item \textbf{CHSEL~\cite{Zhong-RSS-23}:} CHSEL adds free-space reasoning as well as a Quality Diversity (QD) to gradient-based pose registration. CHSEL thus considers geometric consistency and non-penetration via free points. At each timestep, we sample points from the surface of the environment, transform into the grasp frame, and add as free points in the CHSEL optimization, to ensure the object does not collide. We follow the same procedure as ICP to label extrinsic contact. We run 2 CHSEL iterations per update.
    \item \textbf{SCOPE~\cite{Sipos-RSS-23}:} SCOPE utilizes non-penetration, contact kinematics, and force balance for pose and contact estimation via a dual particle filter method. Object pose particles are scored based on non-penetration and their contacts, determined by a Contact Particle Filter (CPF)~\cite{manuelli2016} associated for each object pose particle to estimate the most likely contacts. We modify SCOPE to fit our setup by removing environment pose filtering and replace the F/T sensing with our learned tactile model. Note, this method does not consider geometric consistency. \updatefin{We run two versions of SCOPE. SCOPE (v1) uses 8 pose and 30 contact particles, with 3 iterations every update. SCOPE (v2) uses 64 pose and 200 contact particles, with 20 iterations every update.}
\end{itemize}}
\update{All methods are initialized with the same procedure as TacGraph for fair comparison.}

\subsection{Pose and Contact Estimation} \label{sec:estimation_experiment}

\update{We test pose and contact estimation performance on the objects shown in Fig.~\ref{fig:objects}. For the environment geometry, we use the geometry shown on the right ATI Gamma F/T sensor in Fig.~\ref{fig:exp_setup}. Each object is grasped from a fixture, randomizing the in-hand grasp pose. We perform a fixed set of three angled pokes into the environment. We get sensor feedback before poking and once during each poke, when contact is made. We run inference for each method under two conditions: Vision + Tactile and Tactile. We apply each method iteratively on each subsequent timestep of data.}

\subsubsection{Metrics} 
\update{We determine the ground truth pose of the object using the fixture (tolerance of $\sim$1mm). We use the Average 3D Distance metric~\cite{bauza2023tac2pose} using 1000 points as our object pose metric.}

\update{For the extrinsic contact, we utilize the F/T sensor mounted under the environment. We compute the ground truth contact by sampling a set of candidate contact points on the environment surface $\vec{C} \in \mathbb{R}^{L\times 3}$. We then identify the ground truth by solving the following minimization:
\begin{equation} \label{eq:gt_contact}
    \vec{c}_t^* = \argmin_{l} ||\vec\tau_t^{FT} - \vec{C}_l \times \vec{f}_t^{FT}||_2
\end{equation}
$\vec f_t^{FT}$ and $\vec \tau_t^{FT}$ are the force and torque at the sensor.  The ground truth force is then $\vec{f}_t^*=\vec{f}_t^{FT}$. We compute the euclidean errors on the contact point and force estimate.}

\subsubsection{Results}

\update{We report our pose estimation results in Table~\ref{tab:pose_est}. When vision is available, ICP can perform well, but its performance is subject to sensor noise. CHSEL can rectify some noise via its non-penetration handling. Neither method works well on tactile-only, as they cannot exploit contact to improve. SCOPE outperforms ICP and CHSEL on tactile only, as it can exploit the contact information, but \updatefin{lack of geometric feedback and noise in the filtering reduce precision}. We find that TacGraph consistently outperforms the baselines across nearly all cases, exploiting contact to inform the object pose.}

\update{Table~\ref{tab:contact_est} and Table~\ref{tab:force_est} show the contact estimation performance. When Vision and Tactile are available, we find that geometric based methods perform comparably to ours. However, when vision is removed the contact estimate quickly drifts due to the pose error. In contrast, our method retains similar overall contact estimate performance. Since all methods utilize the learned tactile force model, force estimates differ only slightly - we can see that TacGraph can improve force estimates, due to improved contact location and torque feedback at the grasp.} 

\updatefin{Fig.~\ref{fig:particle_qual} shows qualitative predictions by TacGraph. In Fig.~\ref{fig:particles_qual} we show the progression of $K$ particles by their weight (i.e., $e^{-H(\cdot)}$). We see that TacGraph uses subsequent contacts to correctly identify the correct local solution, rejecting initial solutions that may be consistent with only the local geometric information.}

\subsection{Peg Insertion} 

\begin{table}
    \centering
    \renewcommand{\arraystretch}{1.2}
    \begin{tabular}{c c c c c c}
    \toprule
    \multirow{2}{*}{Object} & \multicolumn{5}{c}{Tactile-Only} \\
    \cmidrule(lr){2-6}
     & ICP & CHSEL & \makecell{SCOPE \\ (v1)} & \makecell{\updatefin{SCOPE} \\ \updatefin{(v2)}} & TacGraph \\
    \toprule
    \texttt{cylinder} & 4 / 10 & 4 / 10 & 4 / 10 & \updatefin{1 / 10} & 7 / 10 \\
    \texttt{rectangle} & 2 / 10 & 6 / 10 & 0 / 10 & \updatefin{1 / 10} & 6 / 10 \\
    \makecell{\texttt{quarter}\\ \texttt{cylinder}} & 0 / 10 & 1 / 10 & 0 / 10 & \updatefin{0 / 10} & 1 / 10 \\
    % \hline
    \texttt{wrench} & 5 / 10 & 3 / 10 & 3 / 10 & \updatefin{1 / 10} & 9 / 10 \\
    \hline
    Overall & 11 / 40 & 14 / 40 & 7 / 40 & \updatefin{3 / 40} & 23 / 40 \\
    \bottomrule
    \end{tabular}
    \caption{\update{Tactile-Only Peg Insertion Results (Success / Attempt)}}
    \label{tab:peg_insertion}
\end{table}

We evaluate how our proposed methodology aids in a prehensile task requiring precise in-hand pose estimation. We perform a peg insertion task with several objects from Fig.~\ref{fig:objects}. Each fixture for insertion is designed per the object geometry with 3 mm clearance. We perform the same inference procedure outlined in Sec.~\ref{sec:estimation_experiment}, fixing the grasps across run for fair comparison. We then take the final rest pose estimate $\orest^*$ and perform an open-loop insertion. We use a F/T sensor mounted under the insertion fixture to stop and release if a force greater than 5 N is registered. If no force threshold is met, the object is released 5 mm above the floor of the insertion fixture. We run our experiment in the challenging tactile-only setting.

\subsubsection*{Results} \update{The peg insertion success rates are reported across 10 presses per object for four of our objects for a total of 40 trials in Tab.~\ref{tab:peg_insertion}. Our method is able to, from tactile alone, achieve precise open-loop peg insertion, outperforming the baselines across the objects. The \texttt{quarter cylinder} object we found had a noisy depth prediction from our model, as the gelsight is not as sensitive in the normal direction of the sensor, which caused a low success rate. We found ICP and CHSEL could align the overall orientation of the object well in some cases, and hence can succeed despite inaccurate poses,} \updatefin{while SCOPE struggled to achieve accurate orientations leading to many failures.}

\section{Discussion} 
\label{sec:conclusion}

\update{In this work, we proposed TacGraph, a factor-graph based method that can utilize the physical constraints of contact, along with sensory feedback, to resolve object pose and contact simultaneously. While our results demonstrate the value of uniting physical constraints and sensory feedback, the work has several limitations.}

\update{TacGraph is still inherently a local method, which means we are sensitive to initialization. A potential extension to explore is enforcing diversity across particles~\cite{Zhong-RSS-23}. In the experiments, our contact interactions here were manually selected - in future, we would like to utilize our estimator online during contact-rich tasks, as well as exploring action selection to explicitly drive down uncertainty in poses~\cite{zhong2025rumi}.}

\update{Our method has several limiting assumptions: that we know object and environment geometries, that no relative slip occurs, and that contact can be described by a single point. Exploring how we can utilize reconstructed object models could relax our geometry assumption~\cite{bauza2023tac2pose,suresh2024neuralfeels}. Utilizing our proposed TacGraph estimator in a control loop could allow corrective actions to avoid high force/torque interactions, thereby avoiding or limiting slip. Additionally, if one could detect a slip event~\cite{veiga2015slip}, the estimator could be re-initialized. Finally, extending to handle multiple contact points or extended contact geometries would help extend this work to a more rich set of interactions.}

% \section*{Acknowledgments}

%% Use plainnat to work nicely with natbib. 

\bibliographystyle{IEEEtran}
\bibliography{references}

\end{document}